\documentclass[runningheads]{llncs}
\usepackage[T1]{fontenc}
\usepackage{graphicx}
\usepackage{amsmath, amssymb}
\usepackage{subcaption}
\usepackage{mdframed}
\usepackage{algorithm2e}
\usepackage{array}
\usepackage{orcidlink}
\usepackage[table]{xcolor}
\usepackage[numbers,sort&compress,square]{natbib}
\usepackage{caption}
\usepackage{etoolbox}
\AtBeginEnvironment{Bibliography}{\vspace{-8pt}}

\definecolor{NavyHdr}{RGB}{25,60,110}
\definecolor{HdrTxt}{RGB}{255,255,255}
\definecolor{RowA}{RGB}{245,248,253}
\definecolor{BestG}{RGB}{198,239,206}
\definecolor{MidAmb}{RGB}{255,235,156}
\definecolor{WorstR}{RGB}{255,199,206}
\definecolor{SecGray}{RGB}{215,215,215}

\begin{document}

\title{Lightweight 3D LiDAR-Based UAV Tracking:\\
       An Adaptive Extended Kalman Filtering Approach%
\thanks{Accepted at the 19th International Conference on Intelligent Autonomous Systems
(IAS-19). To be published in Intelligent Autonomous Systems 19 Proceedings of
IAS-19, Volume 2, in Networks and Systems, Springer Nature. This is the authors’
preprint version; the final authenticated publication will be available via Springer.}}

\titlerunning{Adaptive LiDAR-Based UAV Tracking}

\author{%
  Nivand Khosravi\inst{1}\,\orcidlink{0009-0004-3749-7249} \and
  Meysam Basiri\inst{1}\,\orcidlink{0000-0002-8456-6284} \and
  Rodrigo Ventura\inst{1}\,\orcidlink{0000-0002-5655-9562}}

\authorrunning{N.~Khosravi et al.}

\institute{%
  Institute for Systems and Robotics (ISR), Instituto Superior T\'{e}cnico,
  Lisbon, Portugal\\
  \email{\{nivand.khosravi, meysam.basiri,
          rodrigo.ventura\}@tecnico.ulisboa.pt}}

\maketitle

\begin{abstract}
Accurate relative positioning is crucial for swarm aerial robotics, enabling
coordinated flight and collision avoidance. While vision-based tracking has been
extensively studied, 3D LiDAR-based methods remain underutilized, despite their
robustness in varying lighting conditions. Existing systems often rely on bulky,
power-intensive sensors, making them impractical for small UAVs with strict
payload and energy constraints.

This paper presents a lightweight LiDAR-based UAV tracking system incorporating
an Adaptive Extended Kalman Filtering (AEKF) framework. Our approach effectively
handles the challenges posed by sparse, noisy, and nonuniform point cloud data
generated by non-repetitive scanning 3D LiDARs, ensuring reliable tracking
while remaining suitable for small drones with strict payload constraints. Unlike
conventional filtering techniques, the proposed method dynamically adjusts noise
covariance matrices using innovation and residual statistics, enhancing the
tracking accuracy under real-world conditions. Additionally, a recovery mechanism
ensures continuity of tracking during temporary detection failures caused by
scattered LiDAR returns or occlusions.

Experimental validation using a Livox Mid-360 LiDAR on a DJI F550 UAV
demonstrates successful UAV tracking in real-world conditions, overcoming sparse
returns and intermittent detection while outperforming standard Kalman
filter-based motion tracking during challenging maneuvers. To the best of our
knowledge, this is the first adaptive noise covariance estimation framework
applied to UAV-to-UAV relative tracking using a non-repetitive scanning LiDAR
on a payload-constrained airborne platform. Results confirm our framework enables
reliable relative positioning in GPS-denied environments without multi-sensor
arrays or ground infrastructure.

\end{abstract}

\section{Introduction}

The ability to reliably detect and track small UAVs is crucial in various
scenarios, including swarm coordination, autonomous interception, and airspace
management~\cite{Sahingoz2019,Li2024}. This task becomes particularly challenging
in GPS-denied or restricted environments, such as indoor spaces, urban canyons,
dense foliage, or areas affected by electromagnetic interference, where absolute
positioning is unreliable. In such scenarios, UAVs must rely on onboard sensors
to track cooperative and non-cooperative aerial agents for safe navigation and
effective coordination. Consequently, robust relative positioning and tracking
are critical for mission-critical applications, including search and rescue,
surveillance, and autonomous drone delivery~\cite{Leuci2024}.

Vision-based tracking systems utilizing deep learning models, such as YOLO with
SORT or Siamese networks, perform well under optimal
conditions~\cite{Zhao2022,Li2024}. However, these methods are significantly
degraded in environments with low light, poor visibility, or adverse weather
conditions~\cite{Sahingoz2019,Matou2019}. LiDAR-based sensing presents a viable
alternative by providing three-dimensional measurements that are independent of
lighting conditions~\cite{Abir2023,Hammer2018}. Despite this advantage,
LiDAR-based UAV tracking remains challenging, particularly when tracking small
agile drones with minimal reflective surfaces. Commonly used tracking methods
exhibit limitations in this context. Vision-based systems often exceed the
computational capacity of lightweight UAV platforms, while standard LiDAR
trackers struggle with sparse point cloud data from small UAVs. Multisensor
fusion techniques, although effective, introduce additional processing overhead,
calibration complexities, and system integration challenges, which may conflict
with the strict payload and energy constraints of small aerial
platforms~\cite{Zhong2024}.

Modern compact LiDAR sensors, such as the Livox Mid-360, utilize non-repetitive
scanning patterns that generate fundamentally different data characteristics
compared to traditional LiDAR systems. These scanning patterns result in
irregular point distributions, significant density variations, and inconsistent
feature visibility between frames~\cite{LivoxTech2023,Catalano2023}. Existing
tracking algorithms, which assume uniform scanning patterns and consistent
measurement quality, are incompatible with processing such data, necessitating
the development of novel tracking approaches tailored to these unique sensor
characteristics.

This paper presents a robust LiDAR-based tracking framework designed to address
these challenges. Our approach focuses on effectively processing non-repetitive
pattern point cloud data for relative localization between small UAVs. Our key
contributions are as follows.

\begin{enumerate}
    \item \textbf{Adaptive Filtering for Variable Data Quality:} A filtering
          method that dynamically adjusts noise parameters based on the
          reliability assessment of real-time measurement, addressing the
          inherent variations in non-repetitive scanning data.

    \item \textbf{Robust Data Association in Irregular Point Clouds:}
          Mahalanobis distance-based validation gating to manage uncertainties
          in non-uniform point clouds, ensuring continuous tracking even in
          cluttered environments.

    \item \textbf{Recovery Mechanism for Intermittent Detections:} A
          specialized strategy that maintains tracking continuity during
          measurement gaps by smoothly reinitializing tracking after temporary
          occlusions or target excursions beyond sensor range.

    \item \textbf{Optimized Clustering for Sparse Data:} An optimized DBSCAN
          clustering algorithm tuned for sparse LiDAR returns, facilitating
          precise UAV segmentation without secondary sensor modalities.
\end{enumerate}

Comprehensive validation in relative positioning scenarios between small UAVs
demonstrates that our framework achieves accurate, reliable tracking without
dependence on absolute coordinate systems, addressing a critical gap in UAV
tracking technology for autonomous operations in GPS-denied environments.

The remainder of this paper is structured as follows.
Section~\ref{sec:system} presents an overview of the complete system architecture.
Section~\ref{sec:related} reviews existing LiDAR-based UAV tracking methodologies.
Section~\ref{sec:methodology} details our adaptive Kalman filtering framework.
Section~\ref{sec:experiments} presents experimental results.
Finally, Section~\ref{sec:conclusion} concludes with key findings and future
directions.

\section{System Overview}
\label{sec:system}

\begin{figure}[!ht]
    \centering
    \includegraphics[width=1.0\linewidth]{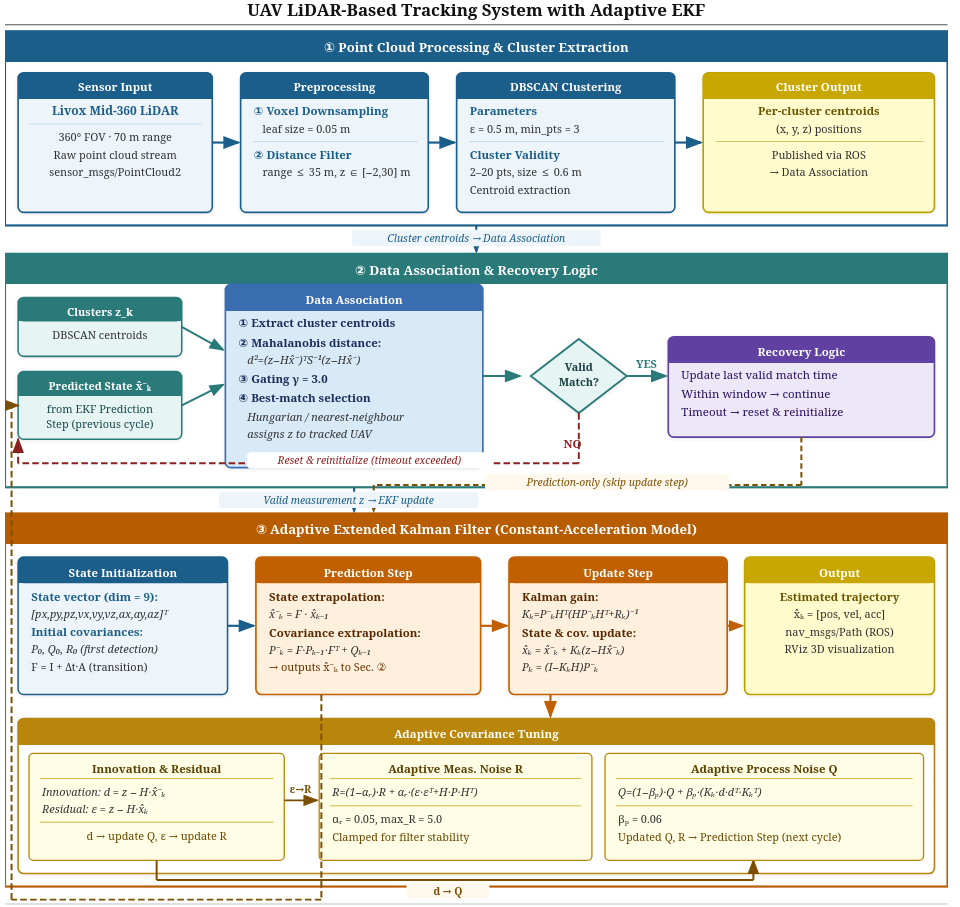}
    \caption{Complete system architecture: LiDAR point cloud processing and
      cluster extraction (Stage~I), data association and recovery logic
      (Stage~II), and Adaptive EKF with online $Q_k$/$R_k$ tuning
      (Stage~III). The feedback arrow from the EKF predicted state $\hat{x}_k^-$
      to the association gate closes the estimation loop.}
    \label{fig:block_diagram}
\end{figure}

Figure~\ref{fig:block_diagram} presents the complete architecture of our
LiDAR-based UAV tracking system. The pipeline is organized into three
interconnected stages: (i)~point cloud processing and DBSCAN-based cluster
extraction, (ii)~data association with Mahalanobis gating and recovery logic,
and (iii)~the Adaptive Extended Kalman Filter with online noise covariance
tuning. Raw scans from the Livox Mid-360 enter Stage~I, where voxel
downsampling, cylindrical ROI filtering, and DBSCAN produce candidate drone
clusters. These clusters feed Stage~II, where they are matched against the
filter's predicted state via Mahalanobis distance gating; unmatched frames
trigger the recovery module. Validated measurements are finally consumed by
Stage~III, where the constant-acceleration EKF propagates the target state and
simultaneously refines both $Q_k$ and $R_k$ from the innovation and residual
statistics.

\section{Related Work}
\label{sec:related}

UAV tracking has been pursued using various sensor modalities, with vision-based
and LiDAR-based systems representing the primary approaches. These methods differ
in detection principles, environmental resilience, computational demands, and
tracking precision.

\subsection{LiDAR-Based UAV Tracking}

LiDAR-based tracking overcomes many limitations of vision-based systems in
challenging conditions such as poor lighting or adverse
weather~\cite{Dogru2022,Abir2023,Catalano2023}. By providing accurate 3D spatial
information, LiDAR sensors enable reliable UAV detection even when visual data
is compromised.

Early work by Razlaw et al.~\cite{Razlaw2019} mounted a Velodyne VLP-16 on a
DJI Matrice 600 MAV and applied Euclidean clustering with a fixed-parameter
Kalman filter to track small \emph{ground-level} objects   not other UAVs  
using fixed noise parameters. Dogru and Marques~\cite{Dogru2022} advanced
ground-based drone detection by combining probabilistic track-before-detect
models with Bayesian filtering on a Velodyne VLP-16, although the method
required substantial computational resources. Catalano et al.~\cite{Catalano2023}
further enhanced the accuracy of tracking in GNSS-denied settings by adaptively
integrating 2--50 solid-state LiDAR scans with specialized Kalman filtering,
though this was limited to field-based ground deployments; the word
``adaptive'' here refers to scan-count selection, not noise covariance
estimation.

Researchers have also explored alternative LiDAR configurations to improve
tracking performance. Gazdag et al.~\cite{Gazdag2024} mounted a Livox AVIA
rosette-scanning LiDAR on a ground pan-tilt turret, combining particle filtering
with OctoMap background subtraction for active tracking, though the bulky
pan-tilt mechanism limits platform portability and noise parameters remain fixed.
Sier et al.~\cite{Sier2023} demonstrated that reinterpreting LiDAR scans as
panoramic images and applying YOLOv5 enables tracking-by-detection from a fixed
ground platform. Abir et al.~\cite{Abir2023} studied point cloud characteristics
of small UAVs using a static Livox Mid-40 array and proposed adaptive
thresholding for detection, but without state estimation. Balla
et al.~\cite{Piotrowski2024} addressed the challenge of discriminating UAVs from
birds in sparse point clouds using attention-based segmentation and CNN
classification on a ground pan-tilt system, focusing on detection and
classification only. Leuci et al.~\cite{Leuci2024} demonstrated camera--LiDAR
fusion in a heterogeneous UAV--UGV team but required multi-sensor calibration
and fixed noise models. Chen et al.~\cite{Chen2016} used a 2D laser scanner on a
quadrotor for collision-free trajectory \emph{generation} in cluttered
environments   a navigation task rather than target tracking.

A structured comparison is provided in Table~\ref{table:LiDAR_approaches}.
Across all existing work, \textbf{no system} simultaneously achieves:
(i)~online adaptive noise covariance estimation ($Q_k$ and $R_k$),
(ii)~UAV-to-UAV relative tracking from an airborne platform, and
(iii)~a non-repetitive scanning LiDAR on a payload-constrained platform.

\begin{table}[!ht]
\centering
\caption{Representative LiDAR-Based UAV Tracking Approaches.}
\label{table:LiDAR_approaches}
\setlength{\tabcolsep}{3pt}
\renewcommand{\arraystretch}{1.15}
\resizebox{\textwidth}{!}{%
\begin{tabular}{|p{2.1cm}|p{2.1cm}|p{2.3cm}|p{2.8cm}|p{3.0cm}|}
\hline
\rowcolor{NavyHdr}
{\color{HdrTxt}\textbf{Approach}} &
{\color{HdrTxt}\textbf{Configuration}} &
{\color{HdrTxt}\textbf{Sensor}} &
{\color{HdrTxt}\textbf{Core Method}} &
{\color{HdrTxt}\textbf{Key Limitation}} \\
\hline
Dogru \& Marques (2022) \cite{Dogru2022}
  & Ground $\!\to\!$ Drone & Velodyne VLP-16
  & Track-before-detect + Bayesian filter
  & High compute; fixed noise model \\
\hline
\rowcolor{RowA}
Catalano et al.\ (2023) \cite{Catalano2023}
  & Ground $\!\to\!$ Drone & Solid-state LiDAR
  & Multi-scan EKF (adaptive scan count only)
  & Ground-only; no noise adaptation \\
\hline
Gazdag et al.\ (2024) \cite{Gazdag2024}
  & Pan-tilt $\!\to\!$ Drone & Livox AVIA (rosette)
  & Particle filter + background subtraction
  & Bulky pan-tilt; fixed noise model \\
\hline
\rowcolor{RowA}
Sier et al.\ (2023) \cite{Sier2023}
  & Fixed ground $\!\to\!$ Drone & Ouster (LiDAR-image)
  & YOLOv5 tracking-by-detection
  & Fixed platform; ML inference cost \\
\hline
Leuci et al.\ (2024) \cite{Leuci2024}
  & UAV+UGV $\!\to\!$ Drone & Camera + LiDAR
  & Multi-sensor fusion + KF
  & Calibration burden; fixed noise \\
\hline
\rowcolor{RowA}
Chen et al.\ (2016) \cite{Chen2016}
  & UAV (self-navigation) & 2D laser + IMU
  & QP trajectory optimisation
  & Navigation only; no target tracking \\
\hline
Razlaw et al.\ (2019) \cite{Razlaw2019}
  & MAV $\!\to\!$ Ground objects & Velodyne VLP-16
  & Euclidean clustering + KF (fixed)
  & Tracks ground targets; fixed noise \\
\hline
\rowcolor{RowA}
Abir et al.\ (2023) \cite{Abir2023}
  & Ground $\!\to\!$ Drone & Livox Mid-40 (static)
  & Adaptive threshold detection
  & Detection only; no state estimation \\
\hline
Balla et al.\ (2024) \cite{Piotrowski2024}
  & Pan-tilt $\!\to\!$ Drone/Bird & Livox AVIA (rosette)
  & CNN classification + segmentation
  & Classification only; no tracking \\
\hline
\textbf{Proposed (AEKF)}
  & \textbf{UAV $\!\to\!$ UAV} & \textbf{Livox Mid-360 (non-rep.)}
  & \textbf{Adaptive EKF ($Q_k$,$R_k$) + DBSCAN + recovery}
  & \textbf{First adaptive noise covariance tracker on airborne LiDAR platform} \\
\hline
\end{tabular}%
}
\end{table}

Existing approaches typically rely on fixed noise parameters in their filtering
frameworks, limiting adaptability to dynamic environments and variable data
quality. Most systems focus on ground-based tracking or require specialised
hardware not suitable for lightweight platforms. Critically, \emph{no existing
work} combines online adaptive noise covariance estimation with UAV-to-UAV
relative tracking using a non-repetitive scanning LiDAR on a payload-constrained
airborne platform.

Our approach differs by implementing an Adaptive Kalman Filter that responds to
real-time variations in both UAV motion and measurement quality from
non-repetitive scanning patterns. As shown in Fig.~\ref{fig:block_diagram}, our
framework integrates point cloud processing, optimised clustering, track
management, and adaptive noise estimation. By dynamically adjusting noise
parameters based on innovation and residual statistics, our system mitigates
tracking errors where measurement quality varies significantly. Comparative
evaluations confirm superior accuracy, robustness, and target reacquisition
performance in challenging conditions with rapid maneuvers and intermittent
detections.

\section{Methodology}
\label{sec:methodology}

We employ an adaptive Kalman filter (AKF) inspired by~\cite{Akhlaghi} that
dynamically adjusts both noise covariances to overcome the inaccuracies of
fixed-parameter filters under varying sensor conditions and dynamic motion.
The framework integrates three components: (i)~point cloud processing and
detection, (ii)~adaptive state estimation, and (iii)~data association and
occlusion handling.

\subsection{LiDAR Data Processing and Object Detection}

This stage extracts potential drone targets from raw LiDAR point clouds through
a three-step process (Algorithm~\ref{alg:lidarprocess}).

\noindent\textbf{Step 1 — Voxel Downsampling.}
The raw cloud $P = \{p_i\}_{i=1}^{N}$, $p_i\in\mathbb{R}^3$, is partitioned
into cubic voxels of side $v = 0.05$\,m; each non-empty voxel $V_j$ is replaced
by its centroid:
\begin{equation}\label{eq:voxel}
c_j = \frac{1}{|V_j|}\sum_{p_i \in V_j} p_i,
\end{equation}
yielding the downsampled cloud $P'$.

\noindent\textbf{Step 2 — Cylindrical ROI Filter.}
Points outside the flight envelope are discarded:
\begin{equation}\label{eq:roi}
P_{\mathrm{ROI}} = \bigl\{ p \in P' \mid
  \sqrt{x^2{+}y^2} < d_{\max},\;\; z_{\min} < z < z_{\max} \bigr\},
\end{equation}
with $d_{\max}=35$\,m and altitude limits $[z_{\min},z_{\max}]$ bounding the
expected flight envelope.

\noindent\textbf{Step 3 — DBSCAN Clustering.}
A point $p \in P_{\mathrm{ROI}}$ is a \emph{core point}~\cite{Ester1996} if
\begin{equation}\label{eq:dbscan_core}
|\mathcal{N}_\epsilon(p)| \geq M_{\min},\quad
\mathcal{N}_\epsilon(p) = \{q \mid \|p{-}q\| \leq \epsilon\}.
\end{equation}
Clusters $\mathcal{C}=\{C_1,\ldots,C_K\}$ are maximal sets of
density-connected points; unreachable points are noise. We use
$\epsilon=0.5$\,m, $M_{\min}=3$, tuned to the 1--4 points the Livox Mid-360
returns per scan on a DJI F550 at 10--35\,m.

\noindent\textbf{Cluster Characterisation and Candidate Validation.}
For each cluster $C_k$, centroid and sample covariance are:
\begin{align}
\boldsymbol{\mu}_k &= \tfrac{1}{|C_k|}\textstyle\sum_{p \in C_k} p, \label{eq:centroid}\\
\Sigma_k &= \tfrac{1}{|C_k|-1}\textstyle\sum_{p \in C_k}
            (p{-}\boldsymbol{\mu}_k)(p{-}\boldsymbol{\mu}_k)^T. \label{eq:cov}
\end{align}
A cluster is a UAV candidate only if
\begin{equation}\label{eq:size_gate}
2 \leq |C_k| \leq N_{\max}, \qquad
\max_i \lambda_i(\Sigma_k) \leq \sigma_{\max}^2,
\end{equation}
where $\lambda_i(\Sigma_k)$ are eigenvalues of $\Sigma_k$. This rejects
single-point spurious returns and large extended objects. The surviving
centroid $\boldsymbol{\mu}_k$ forms measurement $\mathbf{z}_k$.

\begin{mdframed}
\begin{algorithm}[H]
\small
\caption{LiDAR Data Processing and Object Detection}
\label{alg:lidarprocess}
\SetAlgoLined
\KwIn{Raw cloud $P$;\; $v{=}0.05$\,m;\; $d_{\max}{=}35$\,m, $z_{\min}$,
      $z_{\max}$;\; $\epsilon{=}0.5$\,m, $M_{\min}{=}3$;\;
      $N_{\max}$, $\sigma_{\max}$}
\KwOut{Candidate measurement $\mathbf{z}_k$ (or $\varnothing$ if none)}
\tcp{Step 1 — Voxel downsampling (Eq.~\ref{eq:voxel})}
$P' \gets \{\,\mathrm{centroid}(V_j) : V_j \subset P,\;V_j \neq \varnothing\,\}$\;
\tcp{Step 2 — Cylindrical ROI filter (Eq.~\ref{eq:roi})}
$P_{\mathrm{ROI}} \gets \{p{\in}P' \mid \sqrt{x^2{+}y^2}{<}d_{\max},\;
z_{\min}{<}z{<}z_{\max}\}$\;
\tcp{Step 3 — DBSCAN clustering (Eq.~\ref{eq:dbscan_core})}
$\mathcal{C} \gets \mathrm{DBSCAN}(P_{\mathrm{ROI}},\,\epsilon,\,M_{\min})$\;
\tcp{Step 4 — Cluster validation (Eqs.~\ref{eq:centroid}--\ref{eq:size_gate})}
\For{each $C_k \in \mathcal{C}$}{
  Compute $\boldsymbol{\mu}_k$, $\Sigma_k$\;
  \If{$2 \leq |C_k| \leq N_{\max}$ \textbf{and}
      $\max_i \lambda_i(\Sigma_k) \leq \sigma_{\max}^2$}{
    \Return $\mathbf{z}_k = \boldsymbol{\mu}_k$\;
  }
}
\Return $\varnothing$
\end{algorithm}
\end{mdframed}

\subsection{Adaptive Kalman Filtering for UAV State Estimation}

\noindent\textbf{State Space Model.}
A constant-acceleration (CA) model captures agile flight dynamics. The
nine-dimensional state is
\begin{equation}\label{eq:state_vector}
\mathbf{x}_k = \begin{bmatrix}
  p_x & p_y & p_z & v_x & v_y & v_z & a_x & a_y & a_z
\end{bmatrix}^T,
\end{equation}
with transition matrix $F\in\mathbb{R}^{9\times9}$ derived from CA kinematics
at interval $\Delta t$:
\begin{equation}\label{eq:F_matrix}
F = \begin{bmatrix}
  I_3 & \Delta t\,I_3 & \tfrac{\Delta t^2}{2}\,I_3 \\
  0_3 & I_3           & \Delta t\,I_3 \\
  0_3 & 0_3           & I_3
\end{bmatrix}.
\end{equation}
State evolution and LiDAR position measurement are:
\begin{align}
\mathbf{x}_k &= F\,\mathbf{x}_{k-1} + \mathbf{w}_k,\quad
               \mathbf{w}_k \sim \mathcal{N}(\mathbf{0},Q_k), \label{eq:state_evolution}\\
\mathbf{z}_k &= H\,\mathbf{x}_k + \mathbf{v}_k,\quad
               \mathbf{v}_k \sim \mathcal{N}(\mathbf{0},R_k), \label{eq:measurement_equation}
\end{align}
where $H = [I_{3\times3}\ \ 0_{3\times6}]$ selects position components.

\noindent\textbf{Filter Equations.}
The filter cycles through prediction and update at each step~$k$.
\emph{Prediction:}
\begin{align}
\hat{\mathbf{x}}_k^- &= F\,\hat{\mathbf{x}}_{k-1}^+, \label{eq:predict_x}\\
P_k^- &= F\,P_{k-1}^+\,F^T + Q_k. \label{eq:predict_P}
\end{align}
\emph{Update} (given valid $\mathbf{z}_k$):
\begin{align}
S_k &= H\,P_k^-\,H^T + R_k, \label{eq:innov_cov}\\
K_k &= P_k^-\,H^T S_k^{-1}, \label{eq:kalman_gain}\\
\hat{\mathbf{x}}_k^+ &= \hat{\mathbf{x}}_k^- +
                        K_k(\mathbf{z}_k - H\hat{\mathbf{x}}_k^-),
                        \label{eq:update_x}\\
P_k^+ &= (I - K_k H)\,P_k^-. \label{eq:update_P}
\end{align}

\noindent\textbf{Process Noise Adaptation.}
The innovation $\mathbf{d}_k = \mathbf{z}_k - H\hat{\mathbf{x}}_k^-$ grows
when the target accelerates beyond the CA model's prediction. This drives
\begin{equation}\label{eq:Q_update}
Q_k = \alpha\,Q_{k-1} + (1{-}\alpha)\,K_k\,\mathbf{d}_k\mathbf{d}_k^T K_k^T,
\end{equation}
where $\alpha\in(0,1)$ is a forgetting factor. A large $\|\mathbf{d}_k\|$
increases $Q_k \to$ larger $P_k^-$ \eqref{eq:predict_P} $\to$ larger $K_k$
\eqref{eq:kalman_gain}, making the filter more responsive; small innovations
reduce $Q_k$, suppressing noise amplification during smooth flight.

\noindent\textbf{Measurement Noise Adaptation.}
The post-update residual $\boldsymbol{\varepsilon}_k = \mathbf{z}_k -
H\hat{\mathbf{x}}_k^+$ quantifies measurement quality. Elevated
$\|\boldsymbol{\varepsilon}_k\|$ (e.g.\ when the target returns only 1--2
points) triggers
\begin{equation}\label{eq:R_update}
R_k = \beta\,R_{k-1} + (1{-}\beta)\,
      \bigl(\boldsymbol{\varepsilon}_k\boldsymbol{\varepsilon}_k^T
      + H P_k^- H^T\bigr),
\end{equation}
where the $HP_k^-H^T$ term debiases the estimator by removing the prediction
uncertainty contribution. Together, Eqs.~\eqref{eq:Q_update}
and~\eqref{eq:R_update} form a dual-channel self-tuning loop operating purely
from observable statistics. Both $Q_k$ and $R_k$ remain positive semi-definite
for all $k$ by construction, guaranteeing numerical stability.

\subsection{Data Association and Occlusion Handling}

\noindent\textbf{Mahalanobis Gating.}
A candidate $\mathbf{z}_k$ is accepted only if
\begin{equation}\label{eq:mahalanobis_distance}
D_M^2 = (\mathbf{z}_k - H\hat{\mathbf{x}}_k^-)^T S_k^{-1}
         (\mathbf{z}_k - H\hat{\mathbf{x}}_k^-) < \tau,
\end{equation}
where $\tau = \chi^2_{3,\gamma}$ (e.g.\ $\tau\approx14.16$ for $\gamma=0.997$).
The ellipsoidal gate scales automatically with $P_k^-$ and $R_k$: tighter when
certain, wider after measurement gaps.

\noindent\textbf{Occlusion and Reinitialization.}
When no measurement passes the gate, the filter propagates in prediction-only
mode:
\begin{equation}\label{eq:occlusion}
\hat{\mathbf{x}}_k^+ = \hat{\mathbf{x}}_k^-, \qquad P_k^+ = P_k^-.
\end{equation}
$P_k^+$ grows via \eqref{eq:predict_P}, reflecting accumulating uncertainty
while $Q_k$ is frozen. After $t_{\mathrm{occ}}$ consecutive missed steps the
track is reinitialized from the next valid cluster, bounding the prediction
horizon and preventing the unbounded acceleration extrapolation that causes
Fixed CA-KF divergence.

\section{Experimental Results}
\label{sec:experiments}

\begin{figure}[!ht]
    \centering
    \includegraphics[width=0.98\linewidth]{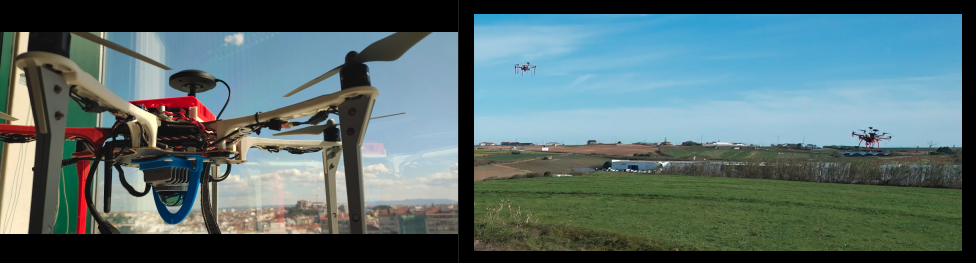}
    \caption{Flight test scenario showing both UAVs in operation. The observer
      UAV (left) is equipped with a Livox Mid-360 LiDAR, while the target UAV
      (right) performs randomized maneuvers for tracking evaluation across
      diverse ranges and orientations.}
    \label{fig:uav_flightrealworld}
\end{figure}

A DJI F550 UAV with a Jetson Nano and Livox Mid-360 LiDAR served as the
observer; a second DJI F550 with an Intel NUC acted as the freely maneuvering
target. RTK positioning on both UAVs provided ground truth, synchronized and
transformed to a common reference frame (Fig.~\ref{fig:uav_flightrealworld}).

\subsection{LiDAR Processing Pipeline}

\begin{figure}[!ht]
  \centering
  \includegraphics[width=\textwidth]{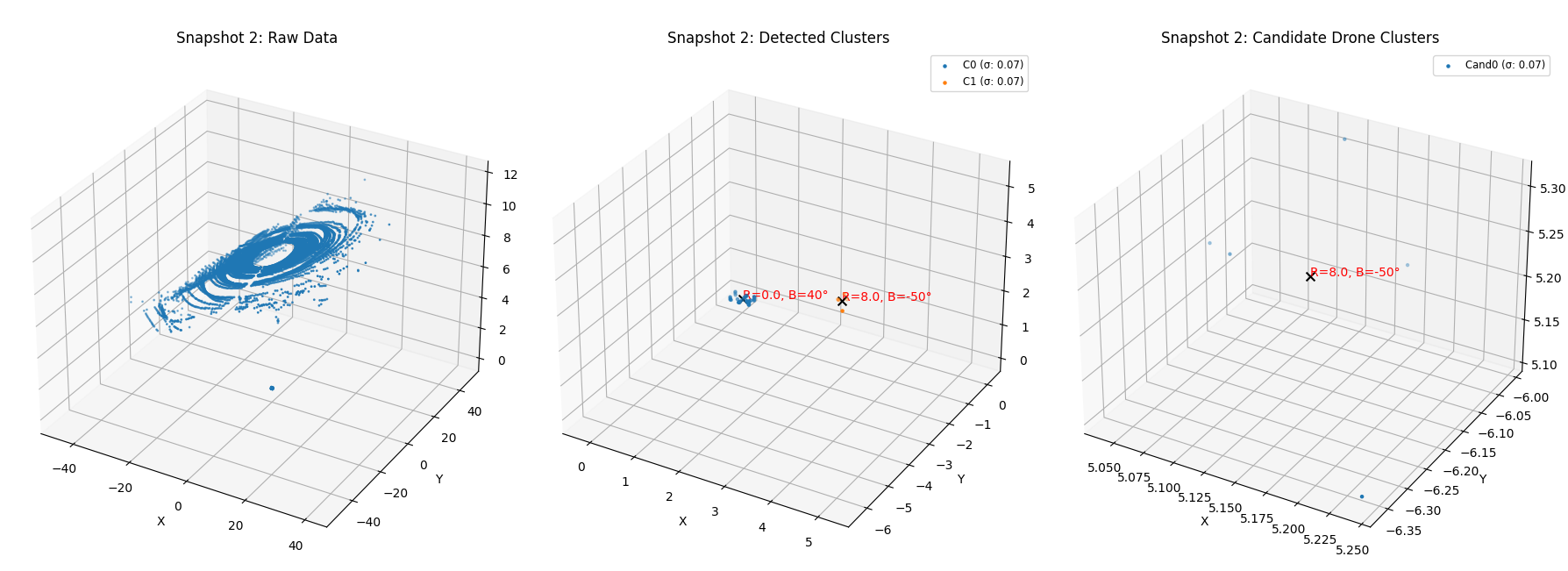}
  \caption{Sequential stages of the LiDAR preprocessing and target-detection
    pipeline. From left to right: raw downsampled point cloud, Euclidean
    clustering with cluster statistics, and final drone candidate selection
    after geometric validation and Mahalanobis gating.}
  \label{fig:preprocessing_and_clustering}
\end{figure}

Figure~\ref{fig:preprocessing_and_clustering} shows the three pipeline stages
on a representative frame: \textbf{(left)} voxel-downsampled cloud;
\textbf{(center)} DBSCAN-clustered segments with annotated centroid and
covariance $\sigma$; \textbf{(right)} selected drone candidate (black cross)
after Mahalanobis gating. In ambiguous cases (birds, foliage producing
drone-like returns), the size-and-extent gate (Eq.~\ref{eq:size_gate}) and the
Mahalanobis gate jointly reject outliers.

\subsection{Results}

To evaluate different filtering strategies we implemented three approaches:
\begin{enumerate}
    \item \textbf{Fixed CA-KF:} A standard constant acceleration Kalman filter
          with fixed (non-adaptive) noise covariance matrices.
    \item \textbf{Particle Filter (PF):} A non-parametric approach using 2000
          particles with systematic resampling.
    \item \textbf{Adaptive CA-EKF (CAEKF):} Our proposed approach, which uses
          the same constant acceleration model as the Fixed CA-KF but
          incorporates adaptive noise estimation and recovery logic.
\end{enumerate}

\begin{figure}[!ht]
\centering
\includegraphics[width=\textwidth]{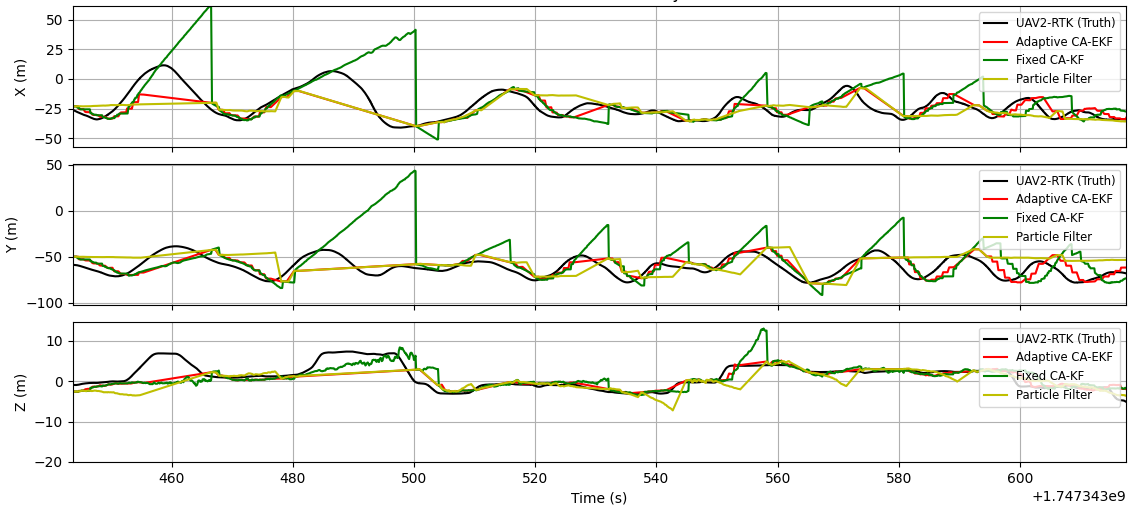}
\caption{Time-series comparison of UAV position along X, Y, and Z axes. RTK
  ground truth (black), Adaptive CA-EKF / CAEKF (red), Fixed CA-KF (green),
  and Particle Filter (yellow). The CAEKF maintains high tracking fidelity,
  while the Fixed CA-KF diverges during measurement gaps
  ($t \approx 480$\,s, $500$\,s, $540$\,s) and the Particle Filter shows
  increased jitter during dynamic maneuvers.}
\label{fig:position_plots}
\end{figure}

Qualitative examination of trajectory estimates (Fig.~\ref{fig:position_plots})
reveals different performance characteristics for each filtering approach. The
CAEKF (red line) demonstrates exceptional tracking fidelity, maintaining
consistent proximity to the RTK reference (black line) throughout the flight
while producing a remarkably smooth trajectory with minimal lag or overshoot
even during aggressive maneuvers.

By contrast, the Fixed CA-KF (green line) exhibits alarming periodic divergence
reaching up to 50\,m from ground truth at timestamps of approximately 480\,s,
500\,s, and 540\,s. These divergences occur precisely during periods when the
target UAV moved beyond the LiDAR's effective range or became too small for
reliable detection during fast horizontal sweeps. Without adaptive covariance
tuning and track recovery, the fixed filter extrapolates the trajectory
indefinitely using the last known acceleration, leading to catastrophic
divergence that renders it unsuitable for high-agility UAV tracking.

The Particle Filter (yellow line) delivers intermediate performance, avoiding
the catastrophic divergence of the Fixed CA-KF but exhibiting persistent jitter
and reduced trajectory smoothness, particularly on the Y-axis between
520--540\,s. This is consistent with well-known sample impoverishment during
rapid state transitions, despite our implementation using 2000 particles.

Table~\ref{tab:performance} presents a detailed comparison of RMSE and
computational metrics. The CAEKF achieves the lowest 3D RMSE (2.8\,m),
representing a \textbf{49.1\%} reduction over the Particle Filter (5.5\,m)
and \textbf{78.5\%} over the Fixed CA-KF (13.0\,m), with X and Y RMSE of
3.0\,m and 4.0\,m versus 6.0/8.0\,m (PF) and 15.0/20.0\,m (Fixed CA-KF).

\begin{table}[!ht]
  \centering
  \caption{Tracking accuracy and computational metrics on Jetson Nano
    under aggressive-maneuver flight. $\dagger$~Peak values elevated only
    during transient adaptation bursts, within four-core capacity.
    $\ddagger$~Lower valid-measurement rate reflects stricter adaptive
    gating   a design strength, not a limitation.}
  \label{tab:performance}
  \renewcommand{\arraystretch}{1.1}
  \setlength{\tabcolsep}{4pt}
  \begin{tabular}{|l|c|c|c|}
    \hline
    \rowcolor{NavyHdr}
    {\color{HdrTxt}\textbf{Metric (Unit)}} &
    {\color{HdrTxt}\textbf{Fixed CA-KF}} &
    {\color{HdrTxt}\textbf{Particle Filter}} &
    {\color{HdrTxt}\textbf{CAEKF (Ours)}} \\
    \hline
    \rowcolor{SecGray}
    \multicolumn{4}{|l|}{%
      \small\textit{(A)~~Tracking Accuracy   RTK ground truth}} \\
    \hline
    RMSE$_X$ (m)    & 15.0 & 6.0 & \textbf{3.0} \\
    \hline
    RMSE$_Y$ (m)    & 20.0 & 8.0 & \textbf{4.0} \\
    \hline
    RMSE$_Z$ (m)    & 4.0  & 2.5 & \textbf{1.5} \\
    \hline
    RMSE$_{3D}$ (m) & 13.0 & 5.5 & \textbf{2.8} \\
    \hline
    \rowcolor{SecGray}
    \multicolumn{4}{|l|}{%
      \small\textit{(B)~~Computational Performance   Jetson Nano (4-core ARM)}} \\
    \hline
    Avg.\ CPU (\%)              & 106.4 & \textbf{0.99}  & 106.5 \\
    \hline
    Total Runtime (s)           & 178.9 & 179.1          & \textbf{174.8} \\
    \hline
    Avg.\ Proc.\ Time (ms)      & 108.5 & \textbf{101.3} & 103.8 \\
    \hline
    \rowcolor{SecGray}
    \multicolumn{4}{|l|}{%
      \small\textit{(C)~~Filter Behaviour}} \\
    \hline
    Valid Measurements (\%)$^{\ddagger}$ & 14.7 & 15.8 & \textbf{8.5} \\
    \hline
    Callback Rate (Hz)          & 8.5  & 9.2            & \textbf{9.3} \\
    \hline
  \end{tabular}
\end{table}

A critical consideration is computational efficiency. The CAEKF maintains an
average processing time of 103.8\,ms per update   lower than the Fixed CA-KF's
108.5\,ms and only marginally above the Particle Filter's 101.3\,ms   while
achieving the highest callback rate (9.3\,Hz) and comparable average CPU usage
(106.5\%). The lower valid-measurement acceptance rate (8.5\% vs.\ 14.7\% and
15.8\%) reflects stricter Mahalanobis gating enabled by adaptive $R_k$,
focusing the filter on high-confidence LiDAR returns. Whereas the Fixed CA-KF
diverges catastrophically in prediction-only mode, the CAEKF dynamically
increases $Q_k$ to bound extrapolation uncertainty until sensor updates resume.
\section{Conclusion}
\label{sec:conclusion}

This paper presented a LiDAR-based UAV tracking framework that uses an Adaptive
Extended Kalman Filter to improve state estimation in dynamic, uncertain
environments. Unlike baseline methods that rely on fixed noise covariance
parameters, our approach adjusts both process and measurement noise in real time
based on innovation and residual statistics. This adaptive strategy improves
tracking precision under a wide range of conditions, as confirmed by real-world
flight experiments that demonstrated improved accuracy, robustness to noise
variations, and responsiveness to rapid UAV maneuvers.

A central feature of our framework is the use of a constant acceleration motion
model, which offers increased flexibility over traditional constant-velocity
models when tracking UAVs performing nonlinear maneuvers. By incorporating
acceleration into the model, the filter is better able to capture sudden changes
in speed and direction   an essential quality for real-world applications where
abrupt movements are common. Our experimental results show that the adaptive
filter significantly reduces estimation errors (RMSE$_{3D}$ of 2.8\,m vs.\
5.5\,m and 13.0\,m for competing methods), prevents divergence, and maintains
stable trajectories compared to standard filtering techniques. The integration
of Mahalanobis distance-based gating enhances measurement association, while our
adaptive noise regulation minimizes tracking errors and avoids overconfidence in
uncertain measurements. Additionally, our track management module successfully
maintains continuous tracking during temporary target loss, effectively
addressing challenges from sparse LiDAR returns and occlusions.

Future work will extend the framework to multi-target tracking in cluttered
environments, improving data association, reducing false positives, and
optimizing computational efficiency. We also plan to investigate the integration
of deep learning-based detection with adaptive filtering to further enhance
robustness in complex aerial operations.

In conclusion, our findings demonstrate the promise of adaptive Kalman filtering
for UAV trajectory estimation and its potential to support scalable, autonomous
aerial systems, particularly in GPS-denied or otherwise challenging environments.

\section*{Acknowledgments}

This work was supported by the Aero.Next project (PRR - C645727867-00000066)
and LARSyS funding (DOI: 10.54499/LA/P/0083/2020, 10.54499/UIDP/50009/2020
and 10.54499/UIDB/50009/2020).

\renewcommand{\bibname}{References}
\begingroup
\let\clearpage\relax

\endgroup


\begin{thebibliography}{99}

\bibitem{Sahingoz2019}
Shi, X., Yang, C., Xie, J., Liang, C., Chen, J.: Anti-Drone System with
Multiple Surveillance Technologies: Architecture, Implementation, and
Challenges. IEEE Communications Magazine \textbf{56}(4), 68--74 (2018).
\doi{10.1109/MCOM.2018.1700430}

\bibitem{Li2024}
Wang, B., Li, Q., Mao, Q., Wang, J., Chen, C.L.P., Shangguan, A., Zhang, H.:
A Survey on Vision-Based Anti Unmanned Aerial Vehicles Methods.
Drones \textbf{8}(9), 518 (2024).
\doi{10.3390/drones8090518}

\bibitem{Leuci2024}
Leuci, C., Dogru, S., Marques, L.: Camera Lidar Fusion for Unmanned Aerial
Vehicle Detection. In: 2024 7th Iberian Robotics Conference (ROBOT),
pp. 1--6 (2024).
\doi{10.1109/ROBOT61475.2024.10796952}

\bibitem{Zhao2022}
Zhu, P., Wen, L., Du, D., Bian, X., Fan, H., Hu, Q., Ling, H.: Detection and
Tracking Meet Drones Challenge. IEEE Transactions on Pattern Analysis and
Machine Intelligence \textbf{44}(11), 7380--7399 (2022).
\doi{10.1109/TPAMI.2021.3119563}

\bibitem{Matou2019}
Vrba, M., He\v{r}t, D., Saska, M.: Onboard Marker-Less Detection and
Localization of Non-Cooperating Drones for Their Safe Interception by an
Autonomous Aerial System. IEEE Robotics and Automation Letters \textbf{4}(4),
3402--3409 (2019)

\bibitem{Abir2023}
Abir, T.A., Kuantama, E., Han, R., Dawes, J., Mildren, R., Nguyen, P.:
Towards Robust Lidar-based 3D Detection and Tracking of UAVs. In: Proceedings
of the Ninth Workshop on Micro Aerial Vehicle Networks, Systems, and
Applications (DroNet '23), pp. 1--7. ACM, New York (2023).
\doi{10.1145/3597060.3597236}

\bibitem{Hammer2018}
Hammer, M., Hebel, M., Laurenzis, M., Arens, M.: Lidar-based detection and
tracking of small UAVs. In: Security + Defence (2018)

\bibitem{Zhong2024}
Nie, C., Ju, Z., Sun, Z., Zhang, H.: 3D Object Detection and Tracking Based
on Lidar-Camera Fusion and IMM-UKF Algorithm Towards Highway Driving. IEEE
Transactions on Emerging Topics in Computational Intelligence \textbf{7}(4),
1242--1252 (2023).
\doi{10.1109/TETCI.2023.3259441}

\bibitem{LivoxTech2023}
Livox Technology Ltd: Livox Mid-360 User Manual. Livox Technology Official
Documentation (2023). \url{https://www.livoxtech.com/mid-360}

\bibitem{Catalano2023}
Catalano, I., Sier, H., Yu, X., Pe\~{n}a Queralta, J., Westerlund, T.: UAV
Tracking with Solid-State Lidars: Dynamic Multi-Frequency Scan Integration.
arXiv preprint arXiv:2304.12125 (2023).
\doi{10.48550/arXiv.2304.12125}

\bibitem{Dogru2022}
Dogru, S., Marques, L.: Drone Detection Using Sparse Lidar Measurements. IEEE
Robotics and Automation Letters \textbf{7}(2), 3062--3069 (2022).
\doi{10.1109/LRA.2022.3145498}

\bibitem{Razlaw2019}
Razlaw, J., Quenzel, J., Behnke, S.: Detection and Tracking of Small Objects
in Sparse 3D Laser Range Data. In: 2019 International Conference on Robotics
and Automation (ICRA), pp. 2967--2973 (2019)

\bibitem{Gazdag2024}
Gazdag, S., Moller, T., Keszler, A., Majdik, A.: Detection and Tracking of
MAVs Using a Rosette Scanning Pattern LiDAR. arXiv:2408.08555 (2024)

\bibitem{Sier2023}
Sier, H., Yu, X., Catalano, I., Queralta, J.P., Zou, Z., Westerlund, T.: UAV
Tracking with Lidar as a Camera Sensor in GNSS-Denied Environments. In: 2023
International Conference on Localization and GNSS (ICL-GNSS), pp. 1--7 (2023).
\doi{10.1109/ICL-GNSS57829.2023.10148919}

\bibitem{Piotrowski2024}
Balla, K., Keszler, A., Gazdag, S., Szir\'{a}nyi, T., Majdik, A.L.:
Detection and Classification of Small-sized UAVs and Birds in Sparse LiDAR
Point Cloud. In: 2024 IEEE International Symposium on Safety Security Rescue
Robotics (SSRR), pp. 172--177 (2024).
\doi{10.1109/SSRR62954.2024.10770035}

\bibitem{Chen2016}
Chen, J., Liu, T., Shen, S.: Online generation of collision-free trajectories
for quadrotor flight in unknown cluttered environments. In: 2016 IEEE
International Conference on Robotics and Automation (ICRA), pp. 1476--1483
(2016).
\doi{10.1109/ICRA.2016.7487283}

\bibitem{Akhlaghi}
Akhlaghi, S., Zhou, N., Huang, Z.: Adaptive adjustment of noise covariance in
Kalman filter for dynamic state estimation. In: 2017 IEEE Power \& Energy
Society General Meeting, pp. 1--5 (2017).
\doi{10.1109/PESGM.2017.8273755}

\bibitem{Ester1996}
Ester, M., Kriegel, H.-P., Sander, J., Xu, X.: A density-based algorithm for
discovering clusters in large spatial databases with noise. In: Proceedings of
the 2nd International Conference on Knowledge Discovery and Data Mining
(KDD-96), pp. 226--231 (1996)

\end{thebibliography}
\end{document}